\documentclass[twocolumn]{article}
\usepackage{epsf}
\renewenvironment{abstract}{\centerline{\large\bf
Abstract}\vspace{0.5ex}\small}{\par\vskip 1ex}
\newenvironment{keywords}{\centerline{\large\bf
Keywords}\vspace{0.5ex}\small}{\par\vskip 1ex}

\def\text{\mbox}

\begin{document}

\title{\vskip -6mm\bf\Large\hrule height5pt \vskip 6mm
Robust Feature Selection by Mutual Information Distributions
\vskip 6mm \hrule height2pt \vskip 5mm}
\author{{\bf Marco Zaffalon} and {\bf Marcus Hutter}\\[3mm]
IDSIA, Galleria 2, CH-6928\ Manno-Lugano, Switzerland\\
\{zaffalon,marcus\}@idsia.ch}
\date{3 June 2002}
\maketitle

\begin{abstract}
Mutual information is widely used in artificial intelligence, in a
descriptive way, to measure the stochastic dependence of discrete
random variables. In order to address questions such as the
reliability of the empirical value, one must consider
sample-to-population inferential approaches. This paper deals with
the distribution of mutual information, as obtained in a Bayesian
framework by a second-order Dirichlet prior distribution. The
exact analytical expression for the mean and an analytical
approximation of the variance are reported. Asymptotic
approximations of the distribution are proposed. The results are
applied to the problem of selecting features for incremental
learning and classification of the naive Bayes classifier. A fast,
newly defined method is shown to outperform the traditional
approach based on empirical mutual information on a number of real
data sets. Finally, a theoretical development is reported that
allows one to efficiently extend the above methods to incomplete
samples in an easy and effective way.
\end{abstract}

\vspace{3mm}


\section{INTRODUCTION}

\label{secInt}
%%%%%%%%%%%%%%%%%%%%%%%%%%%%%%%%%%%%%%%%%%%%%%%%%%%%%%%%%%%%%%%

The {\em mutual information} $I$ (also called {\em cross entropy} or {\em %
information gain}) is a widely used information-theoretic measure for the
stochastic dependency of discrete random variables \cite{Kul68,Cover:91,Soofi:00}.
It is used, for instance, in learning {\em %
Bayesian nets} \cite{ChowLiu68,Pearl88,Buntine:96,Heckerman:98}, where
stochastically dependent nodes shall be connected; it is used to induce
classification trees \cite{Quinlan93}. It is also used to select {\em %
features }for classification problems \cite{DuHaSt01}, i.e.\ to select a
subset of variables by which to predict the {\em class} variable. This is
done in the context of a {\em filter approach }that discards irrelevant
features on the basis of low values of mutual information with the class{\em %
\ }\cite{Lew92,BluLan97,CheHatHayKroMorPagSes02}.

The mutual information (see the definition in Section \ref{DMI})\ can be
computed if the joint chances $\pi _{ij}$ of two random variables $\imath $
and $\jmath $ are known. The usual procedure in the common case of unknown
chances $\pi _{ij}$ is to use the {\em empirical probabilities} $\hat{\pi}%
_{ij}$ (i.e. the sample relative frequencies: $\frac{1}{n}n_{ij}$) as if
they were precisely known chances. This is not always appropriate.
Furthermore, the {\em empirical mutual information} $I\left( {\bf \hat{\pi}}%
\right) $ does not carry information about the reliability of the estimate.
In the Bayesian framework one can address these questions by using a (second
order) prior distribution $p({\bf \pi })$%
%comprising prior information about $\t$.
, which takes account of uncertainty about ${\bf \pi }$. From the prior $p(%
{\bf \pi })$ and the likelihood one can compute the posterior $p({\bf \pi }|%
{\bf n})$, from which the distribution $p(I|{\bf n})$ of the mutual
information can in principle be obtained.

This paper reports, in Section \ref{MR}, the {\em exact} analytical mean of $%
I$ and an analytical $O\left( n^{-3}\right) $-approximation of the variance.
These are reliable and quickly computable expressions following from $p(I|%
{\bf n})$ when a {\em Dirichlet }prior is assumed over ${\bf \pi }$. Such
results allow one to obtain analytical approximations of the distribution of
$I$. We introduce asymptotic approximations of the distribution in Section
\ref{BA}, graphically showing that they are good also for small sample sizes.

The distribution of mutual information is then applied to feature selection.
Section \ref{TPF} proposes two new filters that use {\em credible intervals }%
to robustly estimate mutual information. The filters are empirically tested,
in turn, by coupling them with the {\em naive Bayes classifier }to
incrementally learn from and classify new data. On ten real data sets that
we used, one of the two proposed filters outperforms the traditional filter:
it almost always selects fewer attributes than the traditional one while
always leading to equal or significantly better prediction accuracy of the
classifier (Section \ref{EA}). The new filter is of the same order of
computational complexity as the filter based on empirical mutual
information, so that it appears to be a significant improvement for real
applications.

The proved importance of the distribution of mutual information led us to
extend the mentioned analytical work towards even more effective and
applicable methods. Section \ref{TA} proposes improved analytical
approximations for the tails of the distribution, which are often a critical
point for asymptotic approximations. Section \ref{ETIS} allows the
distribution of mutual information to be computed also from incomplete
samples. Closed-form formulas are developed for the case of feature
selection.

\section{DISTRIBUTION OF MUTUAL INFORMATION}

\label{DMI}

Consider two discrete random variables $\imath $ and $\jmath $ taking values
in $\{1,...,r\}$ and $\{1,...,s\}$, respectively, and an i.i.d.\ random
process with samples $(i,j)\in \{1,...,r\}\times \{1,...,s\}$ drawn with
joint chances $\pi _{ij}$. An important measure of the stochastic dependence
of $\imath $ and $\jmath $ is the mutual information:
\begin{equation}
I({\bf \pi })=\sum_{i=1}^{r}\sum_{j=1}^{s}\pi _{ij}\log \frac{\pi _{ij}}{\pi
_{i+}\pi _{+j}}\text{,}  \label{mi}
\end{equation}
\newline
where $\log $ denotes the natural logarithm and $\pi _{i+}=\sum_{j}\pi _{ij}$
and $\pi _{+j}=\sum_{i}\pi _{ij}$ are marginal chances. Often the chances $%
\pi _{ij}$ are unknown and only a sample is available with $n_{ij}$ outcomes
of pair $(i,j)$. The empirical probability $\hat{\pi}_{ij}=\frac{n_{ij}}{n}$
may be used as a point estimate of $\pi _{ij}$, where $n=\sum_{ij}n_{ij}$ is
the total sample size. This leads to an empirical estimate $I(\hat{{\bf \pi }%
})=\sum_{ij}\frac{n_{ij}}{n}\log \frac{n_{ij}n}{n_{i+}n_{+j}}$ for the
mutual information.

Unfortunately, the point estimation $I(\hat{{\bf \pi }})$ carries no
information about its accuracy. In the Bayesian approach to this problem one
assumes a prior (second order) probability density $p({\bf \pi })$ for the
unknown chances $\pi _{ij}$ on the probability simplex. From this one can
compute the posterior distribution $p({\bf \pi }|{\bf n})\propto p({\bf \pi }%
)\prod_{ij}\pi _{ij}^{n_{ij}}$ (the $n_{ij}$ are multinomially distributed)
and define the posterior probability density of the mutual information:%
\footnote{$I({\bf \pi })$ denotes the mutual information for the specific
chances ${\bf \pi }$, whereas $I$ in the context above is just some
non-negative real number. $I$ will also denote the mutual information {\it %
random variable} in the expectation $E[I]$ and variance $\mbox{Var}[I]$.
Expectations are {\it always} w.r.t.\ to the posterior distribution $p({\bf %
\pi }|{\bf n})$.}
\begin{equation}
p(I|{\bf n})=\int \delta (I({\bf \pi })-I)p({\bf \pi }|{\bf n})d^{rs}{\bf %
\pi .}  \label{midistr}
\end{equation}
\footnote{%
Since $0\leq I({\bf \pi })\leq I_{max}$ with sharp upper bound $I_{max}=\min
\{\log r,\log s\}$, the integral may be restricted to $\int_{0}^{I_{max}}$,
which shows that the domain of $p(I|{\bf n})$ is $[0,I_{max}].$}The $\delta
(\cdot )$ distribution restricts the integral to ${\bf \pi }$ for which $I(%
{\bf \pi })=I$. For large sample size $n\rightarrow \infty $, $p({\bf \pi }|%
{\bf n})$ is strongly peaked around ${\bf \pi }=\hat{{\bf \pi }}$ and $p(I|%
{\bf n})$ gets strongly peaked around the frequency estimate $I=I(\hat{{\bf %
\pi }})$.

\subsection{Results for $I$ under Dirichlet P(oste)riors}

\label{MR}

Many {\em non-informative} priors lead to a Dirichlet posterior distribution
$p({\bf \pi }|{\bf n})\propto \prod_{ij}\pi _{ij}^{n_{ij}-1}$ with
interpretation $n_{ij}=n_{ij}^{\prime }+n_{ij}^{\prime \prime }$, where $%
n_{ij}^{\prime }$ are the number of samples $(i,j)$, and $n_{ij}^{\prime
\prime }$ comprises prior information ($1$ for the uniform prior, ${%
\textstyle{\frac{1}{2}}}$ for Jeffreys' prior, $0$ for Haldane's prior, $%
\frac{1}{rs}$ for Perks' prior \cite{Gelman:95}). In principle this allows
the posterior density $p(I|{\bf n})$ of the mutual information to be
computed.

We focus on the mean $E[I]=\int_{0}^{\infty }Ip(I|{\bf n})\,dI=\int I({\bf %
\pi })p({\bf \pi }|{\bf n})d^{rs}{\bf \pi }$ and the variance $\mbox{Var}%
[I]=E[(I-E[I])^{2}]$. Eq. (\ref{miexex2}) reports the exact mean of the
mutual information:
\begin{eqnarray}
E[I] &=&\frac{{1}}{n}\sum_{ij}n_{ij}[\psi (n_{ij}+1)-\psi (n_{i+}+1)
\nonumber \\
&&-\psi (n_{+j}+1)+\psi (n+1)]\text{,}  \label{miexex2}
\end{eqnarray}
where $\psi $ is the $\psi $-function that for integer arguments is $\psi
(n+1)=-\gamma +\sum_{k=1}^{n}\frac{1}{k}=\log n+O(\frac{1}{n})$, and $\gamma
$ is Euler's constant. The approximate variance is given below:
\begin{eqnarray}
\mbox{Var}[I] &=&\stackrel{O\left( n^{-1}\right) }{\overbrace{\frac{K-J^{2}}{%
n+1}}}+\stackrel{O\left( n^{-2}\right) }{\overbrace{\frac{M+\left(
r-1\right) \left( s-1\right) \left( {\textstyle}\frac{1}{2}-J\right) -Q}{%
\left( n+1\right) \left( n+2\right) }}}  \nonumber \\
&&+O(n^{-3})  \label{varappr}
\end{eqnarray}
where
\begin{eqnarray}
K &=&\sum_{ij}\frac{n_{ij}}{n}\left( \log \frac{n_{ij}n}{n_{i+}n{_{+j}}}%
\right) ^{2},  \nonumber \\
J &=&\sum_{ij}\frac{n_{ij}}{n}\log \frac{n_{ij}n}{n_{i+}n{_{+j}}}=I(\hat{\pi}%
),  \nonumber \\
M &=&\sum_{ij}\left( \frac{1}{n_{ij}}-\frac{1}{n_{i+}}-\frac{1}{n_{+j}}+%
\frac{1}{n}\right) n_{ij}\log \frac{n{_{ij}}n}{n_{i+}n{_{+j}}},  \nonumber \\
Q &=&1-\sum_{ij}\frac{n_{ij}^{2}}{n_{i+}n{_{+j}}}.  \nonumber
\end{eqnarray}

The results are derived in \cite{Hutter:01xentropy}. The result for the mean
was also reported in \cite{WolWol95}, Theorem 10. We are not aware of
similar analytical approximations for the variance. \cite{WolWol95} express
the exact variance as an infinite sum, but this does not allow a
straightforward systematic approximation to be obtained. \cite{Kleiter:99}
used heuristic numerical methods to estimate the mean and the variance.
However, the heuristic estimates are incorrect, as it follows from the
comparison with the analytical results provided here (see \cite{Hutter:01xentropy}).

Let us consider two further points. First, the complexity to compute the
above expressions is of the same order $O(rs)$ as for the empirical mutual
information (\ref{mi}). All quantities needed to compute the mean and the
variance involve double sums only, and the function $\psi $ can be
pre-tabled.

Secondly, let us briefly consider the quality of the approximation of the
variance. The expression for the exact variance has been Taylor-expanded in $%
\left( \frac{rs}{n}\right) $ to produce (\ref{varappr}), so the relative
error ${\frac{\mbox{\scriptsize
Var}[I]_{approx}-\mbox{\scriptsize Var}[I]_{exact}}{\mbox{\scriptsize Var}%
[I]_{exact}}}$ of the approximation is of the order $\left( \frac{rs}{n}%
\right) ^{2}$ , {\em if} $\imath $ and $\jmath $ are dependent. In the
opposite case, the $O\left( n^{-1}\right) $ term in the sum drops itself
down to order $n^{-2}$ resulting in a reduced relative accuracy $O\left(
\frac{rs}{n}\right) $ of (\ref{varappr}). These results were confirmed by
numerical experiments that we realized by Monte Carlo simulation to obtain
``exact'' values of the variance for representative choices of $\pi _{ij}$, $%
r$, $s$, and $n$.

\subsection{Approximating the Distribution}

\label{BA}

Let us now consider approximating the overall distribution of mutual
information based on the formulas for the mean and the variance given in
Section \ref{MR}. Fitting a normal distribution is an obvious possible
choice, as the central limit theorem ensures that $p(I|{\bf n})$ converges
to a Gaussian distribution with mean $E[I]$ and variance $\mbox{Var}[I]$.
Since $I$ is non-negative, it is also worth considering the approximation of
$p(I|{\bf \pi })$ by a Gamma (i.e., a scaled $\chi ^{2}$). Even better, as $%
I $ can be normalized in order to be upper bounded by 1, the Beta
distribution seems to be another natural candidate, being defined for
variables in the $[0,1]$ real interval. Of course the Gamma and the\ Beta
are asymptotically correct, too.
\begin{figure}[tbh]\epsfxsize=86mm
\centerline{\epsfbox{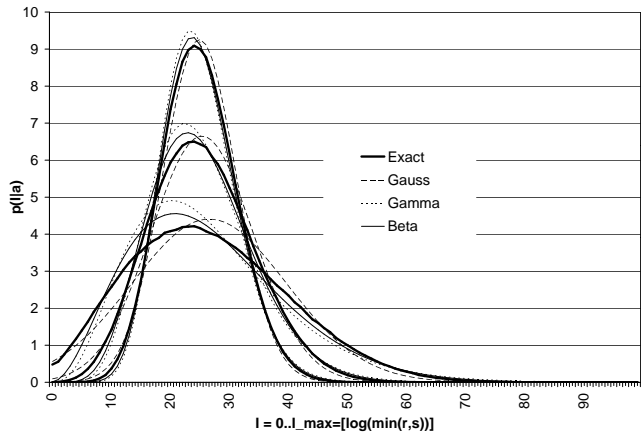}} % [0 0 148 205]
\caption{\label{fig1}\it Distribution of mutual information for
two binary random variables (The labeling of the horizontal axis
is the percentage of I\_$\max $.) There are
three groups of curves, for different choices of counts $%
(n_{11},n_{12},n_{21},n_{22})$. The upper group is related to the vector $%
(40,10,20,80)$, the intermediate one to the vector $(20,5,10,40)$, and the
lower group to $(8,2,4,16)$. Each group shows the ``exact'' distribution and
three approximating curves, based on the Gaussian, Gamma and Beta
distributions.}
\end{figure}

We report a graphical comparison of the different approximations by focusing
on the special case of binary random variables, and on three possible
vectors of counts. Figure \ref{fig1} compares the exact distribution of
mutual information, computed via Monte Carlo simulation, with the
approximating curves. The figure clearly shows that all the approximations
are rather good, with a slight preference for the Beta approximation. The
curves tend to do worse for smaller sample sizes---as it is was expected---.
Higher moments computed in \cite{Hutter:01xentropy} may be used to improve
the accuracy. A method to specifically improve the tail approximation is
given in Section \ref{TA}.

\section{FEATURE SELECTION}

\label{FS}

Classification is one of the most important techniques for knowledge
discovery in databases \cite{DuHaSt01}. A classifier is an algorithm that
allocates new objects to one out of a finite set of previously defined
groups (or {\em classes})\ on the basis of observations on several
characteristics of the objects, called {\em attributes} or {\em features}.
Classifiers can be learnt from data alone, making explicit the knowledge
that is hidden in raw data, and using this knowledge to make predictions
about new data.

Feature selection is a basic step in the process of building classifiers
\cite{BluLan97,DasLiu97,LiuMot98}. In fact, even if theoretically more
features should provide one with better prediction accuracy (i.e., the
relative number of correct predictions), in real cases it has been observed
many times that this is not the case \cite{KolSah96}. This depends on the
limited availability of data in real problems:\ successful models seem to be
in good balance of model complexity and available information. In facts,
feature selection tends to produce models that are simpler, clearer,
computationally less expensive and, moreover, providing often better
prediction accuracy. Two major approaches to feature selection are commonly
used \cite{JohKohPfl94}: {\em filter} and {\em wrapper} models. The filter
approach is a preprocessing step of the classification task. The wrapper
model is computationally heavier, as it implements a search in the feature
space.

\subsection{The Proposed Filters}

\label{TPF}

From now on we focus our attention on the filter approach. We consider the
well-known filter (F)\ that computes the empirical mutual information
between features and the class, and discards low-valued features \cite{Lew92}%
. This is an easy and effective approach that has gained popularity with
time. Cheng reports that it is particularly well suited to jointly work with
Bayesian network classifiers, an approach by which he won the {\em 2001
international knowledge discovery competition} \cite{CheHatHayKroMorPagSes02}%
. The ``Weka'' data mining package implements it as a standard system tool
(see \cite{WitFra99}, p. 294).

A problem with this filter is the variability of the empirical mutual
information with the sample. This may allow wrong judgments of relevance to
be made, as when features are selected by keeping those for which mutual
information exceeds a fixed threshold $\varepsilon .$\ In order for the
selection to be robust, we must have some guarantee about the actual value
of mutual information.

We define two new filters. The {\em backward filter }(BF) discards an
attribute if its value of mutual information with the class is less than or
equal to $\varepsilon $ with given (high)\ probability $p$. The {\em forward
filter} (FF) includes an attribute if the mutual information is greater than
$\varepsilon $ with given (high)\ probability $p$. BF\ is a conservative
filter, because it will only discard features after observing substantial
evidence supporting their irrelevance. FF instead will tend to use fewer
features, i.e. only those for which there is substantial evidence about them
being useful in predicting the class.

The next sections present experimental comparisons of the new filters and
the original filter F.

\section{EXPERIMENTAL ANALYSES}

\label{EA}

For the following experiments we use the naive Bayes classifier \cite{DuHa73}%
. This is a good classification model---despite its simplifying assumptions,
see \cite{DoPa97}---, which often competes successfully with the
state-of-the-art classifiers from the machine learning field, such as C4.5
\cite{Quinlan93}. The experiments focus on the incremental use of the naive
Bayes classifier, a natural learning process when the data are available
sequentially: the data set is read instance by instance; each time, the
chosen filter selects a subset of attributes that the naive Bayes uses to
classify the new instance; the naive Bayes then updates its knowledge by
taking into consideration the new instance and its actual class. The
incremental approach allows us to better highlight the different behaviors
of the empirical filter (F)\ and those based on credible intervals on mutual
information (BF and FF). In fact, for increasing sizes of the learning set
the filters converge to the same behavior.

For each filter, we are interested in experimentally evaluating two
quantities: for each instance of the data set, the average number of correct
predictions (namely, the prediction accuracy) of the naive Bayes classifier
up to such instance; and the average number of attributes used. By these
quantities we can compare the filters and judge their effectiveness.

The implementation details for the following experiments include: using the
Beta approximation (Section \ref{BA}) to the distribution of mutual
information; using the uniform prior for the naive Bayes classifier and all
the filters; using natural logarithms everywhere; and setting the level $p$
of the posterior probability to $0.95$. As far as $\epsilon $ is concerned,
we cannot set it to zero because the probability that two variables are
independent ($I=0$) is zero according to the inferential Bayesian approach.
We can interpret the parameter $\epsilon $ as a degree of dependency
strength below which attributes are deemed irrelevant. We set $\epsilon $ to
$0.003$, in the attempt of only discarding attributes with negligible impact
on predictions. As we will see, such a low threshold can nevertheless bring
to discard many attributes.

\subsection{Data Sets}

\label{DS}

Table \ref{tab1} lists the 10 data sets used in the experiments. These are
real data sets on a number of different domains. For example, Shuttle-small
reports data on diagnosing failures of the space shuttle; Lymphography and
Hypothyroid are medical data sets; Spam is a body of e-mails that can be
spam or non-spam; etc.

\begin{table}\centering
\caption{\it Data sets used in the experiments, together with their
number of features, of instances and the relative frequency of the
majority class. All but the Spam data sets are available from the
UCI repository of machine learning data sets \cite{MurAha95}. The
Spam data set is described in \cite{AndKouKonChaPalSpy00} and
available from Androutsopoulos's web page.\label{tab1}}\medskip
\begin{tabular}{cccc}
\hline
Name & \# feat. & \# inst. & maj. class \\ \hline
\multicolumn{1}{l}{Australian} & \multicolumn{1}{r}{36} & \multicolumn{1}{r}{
690} & \multicolumn{1}{r}{0.555} \\
\multicolumn{1}{l}{Chess} & \multicolumn{1}{r}{36} & \multicolumn{1}{r}{3196}
& \multicolumn{1}{r}{0.520} \\
\multicolumn{1}{l}{Crx} & \multicolumn{1}{r}{15} & \multicolumn{1}{r}{653} &
\multicolumn{1}{r}{0.547} \\
\multicolumn{1}{l}{German-org} & \multicolumn{1}{r}{17} & \multicolumn{1}{r}{
1000} & \multicolumn{1}{r}{0.700} \\
\multicolumn{1}{l}{Hypothyroid} & \multicolumn{1}{r}{23} &
\multicolumn{1}{r}{2238} & \multicolumn{1}{r}{0.942} \\
\multicolumn{1}{l}{Led24} & \multicolumn{1}{r}{24} & \multicolumn{1}{r}{3200}
& \multicolumn{1}{r}{0.105} \\
\multicolumn{1}{l}{Lymphography} & \multicolumn{1}{r}{18} &
\multicolumn{1}{r}{148} & \multicolumn{1}{r}{0.547} \\
\multicolumn{1}{l}{Shuttle-small} & \multicolumn{1}{r}{8} &
\multicolumn{1}{r}{5800} & \multicolumn{1}{r}{0.787} \\
\multicolumn{1}{l}{Spam} & \multicolumn{1}{r}{21611} & \multicolumn{1}{r}{
1101} & \multicolumn{1}{r}{0.563} \\
\multicolumn{1}{l}{Vote} & \multicolumn{1}{r}{16} & \multicolumn{1}{r}{435}
& \multicolumn{1}{r}{0.614} \\ \hline
\end{tabular}
\end{table}

The data sets presenting non-nominal features have been pre-discretized by
MLC++ \cite{KoJoLoMaPf94}, default options. This step may remove some
attributes judging them as irrelevant, so the number of \ features in the
table refers to the data sets after the possible discretization. The
instances with missing values have been discarded, and the third column in
the table refers to the data sets without missing values. Finally, the
instances have been randomly sorted before starting the experiments.

\subsection{Results}

\label{R}

In short, the results show that FF outperforms the commonly used filter F,
which in turn, outperforms the filter BF. FF leads either to the same
prediction accuracy as F or to a better one, using substantially fewer
attributes most of the times. The same holds for F versus BF.

In particular, we used the {\em two-tails paired t test} at level 0.05 to
compare the prediction accuracies of the naive Bayes with different filters,
in the first $k$ instances of the data set, for each $k$.

On eight data sets out of ten, both the differences between FF\ and F, and
the differences between F and BF,\ were never statistically significant,
despite the often-substantial different number of used attributes, as from
Table \ref{tab2}.

\begin{table}\centering
\caption{\it Average number of attributes selected by the filters on
the entire data set, reported in the last three columns. The
second column from left reports the original number of features.
In all but one case, FF\ selected fewer features than F, sometimes
much fewer;\ F usually selected much fewer features than BF, which
was very conservative. Boldface names refer to data sets on which
prediction accuracies where significantly
different.\label{tab2}}\medskip
\begin{tabular}{ccccc}
\hline
Data set & \# feat. & FF & F & BF \\ \hline
\multicolumn{1}{l}{Australian} & \multicolumn{1}{r}{36} & \multicolumn{1}{r}{
32.6} & \multicolumn{1}{r}{34.3} & \multicolumn{1}{r}{35.9} \\
\multicolumn{1}{l}{\bf Chess} & \multicolumn{1}{r}{36} & \multicolumn{1}{r}{
12.6} & \multicolumn{1}{r}{18.1} & \multicolumn{1}{r}{26.1} \\
\multicolumn{1}{l}{Crx} & \multicolumn{1}{r}{15} & \multicolumn{1}{r}{11.9}
& \multicolumn{1}{r}{13.2} & \multicolumn{1}{r}{15.0} \\
\multicolumn{1}{l}{German-org} & \multicolumn{1}{r}{17} & \multicolumn{1}{r}{
5.1} & \multicolumn{1}{r}{8.8} & \multicolumn{1}{r}{15.2} \\
\multicolumn{1}{l}{Hypothyroid} & \multicolumn{1}{r}{23} &
\multicolumn{1}{r}{4.8} & \multicolumn{1}{r}{8.4} & \multicolumn{1}{r}{17.1}
\\
\multicolumn{1}{l}{Led24} & \multicolumn{1}{r}{24} & \multicolumn{1}{r}{13.6}
& \multicolumn{1}{r}{14.0} & \multicolumn{1}{r}{24.0} \\
\multicolumn{1}{l}{Lymphography} & \multicolumn{1}{r}{18} &
\multicolumn{1}{r}{18.0} & \multicolumn{1}{r}{18.0} & \multicolumn{1}{r}{18.0
} \\
\multicolumn{1}{l}{Shuttle-small} & \multicolumn{1}{r}{8} &
\multicolumn{1}{r}{7.1} & \multicolumn{1}{r}{7.7} & \multicolumn{1}{r}{8.0}
\\
\multicolumn{1}{l}{\bf Spam} & \multicolumn{1}{r}{21611} &
\multicolumn{1}{r}{123.1} & \multicolumn{1}{r}{822.0} & \multicolumn{1}{r}{
13127.4} \\
\multicolumn{1}{l}{Vote} & \multicolumn{1}{r}{16} & \multicolumn{1}{r}{14.0}
& \multicolumn{1}{r}{15.2} & \multicolumn{1}{r}{16.0} \\ \hline
\end{tabular}
\end{table}

The remaining cases are described by means of the following figures. Figure
\ref{fig2} shows that FF allowed the naive Bayes to significantly do better
predictions than F for the greatest part of the Chess data set. The maximum
difference in prediction accuracy is obtained at instance 422, where the
accuracies are 0.889 and 0.832 for the cases FF and F, respectively. Figure
\ref{fig2} does not report the BF case, because there is no significant
difference with the F curve. The good performance of FF was obtained using
only about one third of the attributes (Table \ref{tab2}).

\begin{figure}[tbh]\epsfxsize=86mm
\centerline{\epsfbox{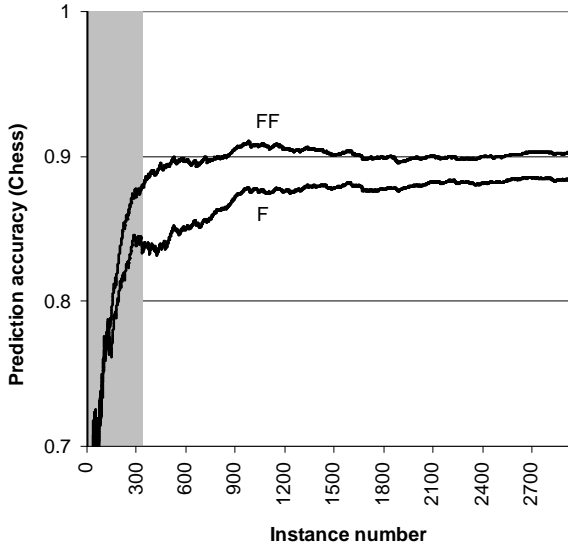}} % [0 0 148 205]
\caption{\label{fig2}\it Comparison of the prediction accuracies of
the naive Bayes with filters F and FF on the Chess data set. The gray area
denotes differences that are not statistically significant.}
\end{figure}

\begin{figure}[tbh]\epsfxsize=86mm
\centerline{\epsfbox{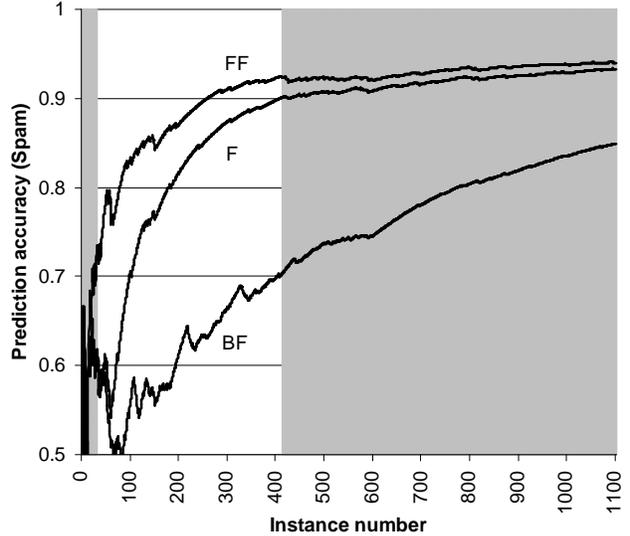}} % [0 0 148 205]
\caption{\label{fig3}\it Prediction accuracies of
the naive Bayes with filters F, FF and BF on the Spam data set. The
differences between F and FF are significant in the range of observations
32--413. The differences between F and BF are significant from observations
65 to the end (this significance is not displayed in the picture).}
\end{figure}

\begin{figure}[tbh]\epsfxsize=86mm
\centerline{\epsfbox{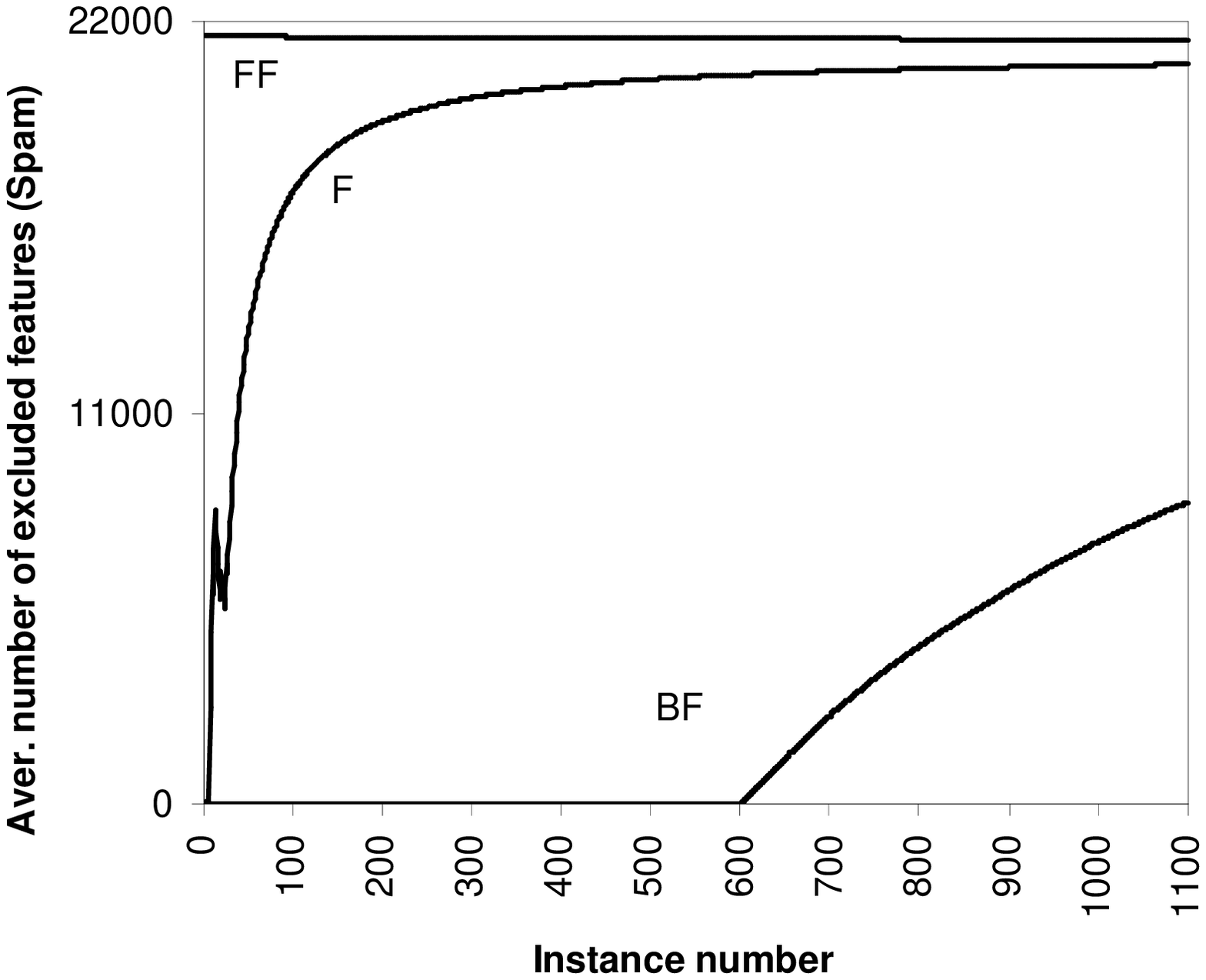}} % [0 0 148 205]
\caption{\label{fig4}\it Comparison of the prediction accuracies of
the naive Bayes with filters F and FF on the Chess data set. The gray area
denotes differences that are not statistically significant.}
\end{figure}

Figure \ref{fig3} compares the accuracies on the Spam data set. The
difference between the cases FF and F is significant in the range of
instances 32--413, with a maximum at instance 59 where accuracies are 0.797
and 0.559 for FF and F, respectively. BF is significantly worse than F from
instance 65 to the end. This excellent performance of FF is even more
valuable considered the very low number of attributes selected for
classification. In the Spam case, attributes are binary and correspond to
the presence or absence of words\ in an e-mail and the goal is to decide
whether or not the e-mail is spam. All the 21611 words found in the body of
e-mails were initially considered. FF shows that only an average of about\
123 relevant words is needed to make good predictions. Worse predictions are
made using F and BF, which select, on average, about 822 and 13127 words,
respectively. Figure \ref{fig4} shows the average number of excluded
features for the three filters on the Spam data set. FF suddenly discards
most of the features, and keeps the number of selected features almost
constant over all the process. The remaining filters tend to such a number,
with different speeds, after initially including many more features than FF.%

In summary, the experimental evidence supports the strategy of only using
the features that are reliably judged as carrying useful information to
predict the class, provided that the judgment can be updated as soon as new
observations are collected. FF almost always selects fewer features than F,
leading to a prediction accuracy at least as good as the one F leads to. The
comparison between F and BF is analogous, so FF appears to be the best
filter and BF the worst. However, the conservative nature of BF might turn
out to be successful when data are available in groups, making the
sequential updating be not viable. In this case, it does not seem safe to
take strong decisions of exclusion that have to be maintained for a number
of new instances, unless there is substantial evidence against the relevance
of an attribute.

\section{EXTENSIONS}

\label{E}

\subsection{Tails Approximation}

\label{TA}

The expansion of $p(I|{\bf n})$ around the mean can be a poor estimate for
extreme values $I\approx 0$ or $I\approx I_{max}$ and it is better to use
tail approximations. The scaling behavior of $p(I|{\bf n})$ can be
determined in the following way: $I({\bf \pi })$ is small iff $\pi _{\imath
\jmath }$ describes near independent random variables $\imath $ and $\jmath $%
. This suggests the reparameterization $\pi _{ij}=\tilde{\pi}_{i{+}}\tilde{%
\pi}_{{+}j}+\Delta _{ij}$ in the integral (\ref{midistr}). Only small $%
\Delta $ can lead to small $I({\bf \pi })$. Hence, for small $I$
we may expand $I({\bf \pi })$ in $\Delta $ in expression
(\ref{midistr}). Correctly taking into account the constraints on
$\Delta $, a scaling argument shows that $p(I|{\bf n})\sim
I^{{\frac{1}{2}}(r-1)(s-1)-1}$. Similarly we get the scaling
behavior of $p(I|{\bf n})$ around $I\approx I_{max}=\min \{\log
r,\log s\}$. $I({\bf \pi })$ can be written as $H(\imath
)-H(\imath |\jmath ) $, where $H$ is the entropy. Without loss of
generality $r\leq s$. If the prior $p({\bf \pi }|{\bf n})$
converges to zero for $\pi _{ij}\rightarrow 0$
sufficiently rapid (which is the case for the Dirichlet for not too small $%
{\bf n}$), then $H(\imath )$ gives the dominant contribution when $%
I\rightarrow I_{max}$. The scaling behavior turns out to be $p(I_{max}-I_{c}|%
{\bf n})\sim I_{c}^{\frac{r-3}{2}}$. These expressions including the
proportionality constants in case of the Dirichlet distribution are derived
in the journal version \cite{HutZaf02_tr}.

\subsection{Incomplete Samples}

\label{ETIS}

In the following we generalize the setup to include the case of missing
data, which often occurs in practice. For instance, observed instances often
consist of several features plus class label, but some features may not be
observed, i.e.\ if $i$ is a feature and $j$ a class label, from the pair $%
(i,j)$ only $j$ is observed. We extend the contingency table
$n_{ij}$ to include $n_{?j}$, which counts the number of instances
in which only the class $j$ is observed (= number of $(?,j)$
instances). It has been shown that using such partially observed
instances can improve classification accuracy \cite{Little:87}. We
make the common assumption that the missing-data mechanism is
ignorable (missing at random and distinct) \cite{Little:87}, i.e.\
the probability distribution of class labels $j$ of
instances with missing feature $i$ is assumed to coincide with the marginal $%
\pi _{{+}j}$.

The probability of a specific data set ${\bf D}$ of size $N=n+n_{{+}?}$ with
contingency table ${\bf N}=\{n_{ij},n_{i?}\}$ given ${\bf \pi }$, hence, is $%
p({\bf D}|{\bf \pi },n,n_{{+}?})=\prod_{ij}\pi _{ij}^{n_{ij}}\prod_{i}\pi _{i%
{+}}^{n_{i?}}$. Assuming a uniform prior $p({\bf \pi })\sim \delta (\pi
_{++}-1)$ Bayes' rule leads to the posterior $p({\bf \pi }|{\bf N})\sim
\prod_{ij}\pi _{ij}^{n_{ij}}\prod_{i}\pi _{i{+}}^{n_{i?}}\delta (\pi
_{++}-1) $. The mean and variance of $I$ in leading order in $N^{-1}$ can be
shown to be
\begin{eqnarray*}
E[{\bf \pi }] &=& I({\bf \hat{\pi}})+O(N^{-1}), \\
\mbox{Var}[I] &=& \frac{1}{N}[\tilde{K}-\tilde{J}^{2}/\tilde{Q}-\tilde{P}%
]+O(N^{-2}),
\end{eqnarray*}
where
\begin{eqnarray*}
\hat{\pi}_{ij} \!\!\!&=&\!\!\! \frac{n_{i+}+n_{i?}}{N}\frac{n_{ij}}{n_{i+}},\ \rho _{ij}=N%
\frac{\hat{\pi}_{ij}^{2}}{n_{ij}},\ \rho _{i?}=N\frac{\hat{\pi}_{i+}^{2}}{%
n_{i?}}, \\
\tilde{Q}_{i?} \!\!\!&=&\!\!\! \frac{\rho _{i?}}{\rho _{i?}+\rho _{i+}},\qquad\qquad\qquad \tilde{Q}%
=\sum_{i}\rho _{i{+}}\tilde{Q}_{i?}, \\
\tilde{K} \!\!\!&=&\!\!\! \sum_{ij}\rho _{ij}\left( \log \frac{\hat{\pi}_{ij}}{\hat{\pi}%
_{i+}\hat{\pi}_{+j}}\right) ^{2},\ \tilde{P}=\sum_{i}\frac{\tilde{J}%
_{i+}^{2}Q_{i?}}{\rho _{i?}}, \\
\tilde{J} \!\!\!&=&\!\!\! \sum_{i}\tilde{J}_{i{+}}\tilde{Q}_{i?},\qquad \tilde{J}_{i{+}%
}=\sum_{j}\rho _{ij}\log \frac{\hat{\pi}_{ij}}{\hat{\pi}_{i+}\hat{\pi}_{+j}}.
\end{eqnarray*}

The derivation will be given in the journal version \cite{HutZaf02_tr}. Note
that for the complete case $n_{i?}=0$, we have $\hat{\pi}_{ij}=\rho _{ij}=%
\frac{n_{ij}}{n}$, $\rho _{i?}=\infty $, $\tilde{Q}_{i?}=1$, $\tilde{J}=J$, $%
\tilde{K}=K$, and $\tilde{P}=0$, consistently with (\ref{varappr}).
Preliminary experiments confirm that FF outperforms F also when feature
values are partially missing.

All expressions involve at most a double sum, hence the overall
computation time is $O(rs)$. For the case of missing class labels,
but no missing features, symmetrical formulas exist. In the
general case of missing features and missing class labels
estimates for ${\bf \hat{\pi}}$ have to be obtained numerically,
e.g.\ by the EM algorithm \cite{Chen:74} in time $O(\#\!\cdot
\!rs)$, where $\#$ is the number of iterations of EM. In
\cite{HutZaf02_tr} we derive a closed form expression for the
covariance of $p({\bf \pi }|{\bf N})$ and the variance of $I$ to
leading order which can be evaluated in time $O(s^{2}\left(
s+r\right) )$. This is reasonably fast, if the number of classes
is small, as is often the case in practice. Note that these
expressions converge for $N\rightarrow \infty $ to the exact
values. The missingness needs not to be small.

\section{CONCLUSIONS}

\label{C}

This paper presented ongoing research on the distribution of
mutual information and its application to the important issue of
feature selection. In the former case, we provide fast analytical
formulations that are shown to approximate the distribution well
also for small sample sizes.\ Extensions are presented that, on
one side, allow improved approximations of the tails of the
distribution to be obtained, and on the other, allow the
distribution to be efficiently approximated also in the common
case of incomplete samples. As far as feature selection is
concerned, we empirically showed that a newly defined filter based
on the distribution of mutual information outperforms the popular
filter based on empirical mutual information. This result is
obtained jointly with the naive Bayes classifier.

More broadly speaking, the presented results are important since
reliable estimates of mutual information can significantly improve
the quality of applications, as for the case of feature selection
reported here. The significance of the results is also enforced by
the many important models based of mutual information. Our results
could be applied, for instance, to {\em robustly }infer
classification trees. Bayesian networks can be inferred by using
credible intervals for mutual information, as proposed by
\cite{Kleiter:99}. The well-known Chow and Liu's approach
\cite{ChowLiu68} to the inference of tree-networks might be
extended to credible intervals (this could be done by joining
results presented here and in past work \cite{Zaf01a}).

Overall, the distribution of mutual information seems to be a
basis on which reliable and effective uncertain models can be
developed.

\subsubsection*{Acknowledgements}

Marcus Hutter was supported by SNF grant 2000-61847.00 to J\"{u}rgen
Schmidhuber.

\begin{small}
\bibliographystyle{alpha}

\newcommand{\etalchar}[1]{$^{#1}$}

\end{small}

\end{document}